\pdfoutput=1

\documentclass[11pt]{article}

\usepackage{EACL2023}

\usepackage{times}
\usepackage{latexsym}
\usepackage{graphicx}
\usepackage{stfloats}
\usepackage{float}
\usepackage{enumitem}
\usepackage{hyperref}

\usepackage{colortbl}
\usepackage{tabularx}
\usepackage[dvipsnames]{xcolor}

\usepackage{listings}
\usepackage{xcolor}
\lstdefinestyle{FlushLeftPrompt}{
    basicstyle=\ttfamily\small, 
    breaklines=true,
    breakatwhitespace=true,
    frame=none,
    columns=fixed,
    numbers=none, 
    postbreak={},
    breakindent=0pt,
}

\usepackage[T1]{fontenc}

\usepackage[utf8]{inputenc}

\usepackage{microtype}

\usepackage{inconsolata}
\usepackage{multirow}

\title{Multimodal Claim Extraction for Fact-Checking}

\author{
  \textbf{
    Joycelyn Teo\textsuperscript{1}\textsuperscript{,}\thanks{\ \ This work was partially done during Joycelyn's research visit at Cambridge.}\ , 
    Rui Cao\textsuperscript{2}, 
    Zhenyun Deng\textsuperscript{2}, 
    Zifeng Ding\textsuperscript{2}
  }\\
  \textbf{
    Michael Sejr Schlichtkrull\textsuperscript{3}, 
    Andreas Vlachos\textsuperscript{2}
  }\\[0.5em]
  \textsuperscript{1}Defence Science and Technology Agency, Singapore\\
  \textsuperscript{2}University of Cambridge, UK, 
  \textsuperscript{3}Queen Mary University of London, UK\\[0.5em]
  \texttt{tlimeijo@dsta.gov.sg, \{rc990,zd302,zd320,av308\}cam.ac.uk}\\
  \texttt{m.schlichtkrull@qmul.ac.uk}
}

\begin{document}
\maketitle
\begin{abstract}
Automated Fact-Checking (AFC) relies on claim extraction as a first step, yet existing methods largely overlook the multimodal nature of today’s misinformation. Social media posts often combine short, informal text with images such as memes, screenshots, and photos, creating challenges that differ from both text-only claim extraction and well-studied multimodal tasks like image captioning or visual question answering.
In this work, we present the first benchmark for multimodal claim extraction from social media, consisting of posts containing text and one or more images, annotated with gold-standard claims derived from real-world fact-checkers.
We evaluate state-of-the-art multimodal LLMs (MLLMs) under a three-part evaluation framework (semantic alignment, faithfulness, and decontextualization) and find that baseline MLLMs struggle to model rhetorical intent and contextual cues. 
To address this, we introduce MICE, an intent-aware framework which shows improvements in intent-critical cases.\footnote{The MMCE dataset is available at \url{https://huggingface.co/datasets/joycelynt/MMCE}, and the code can be accessed at \url{https://github.com/jt9080/MMCE-Paper}.}

\end{abstract}

\section{Introduction}

\begin{figure}[t!]
    \centering
    \includegraphics[width=1\linewidth]{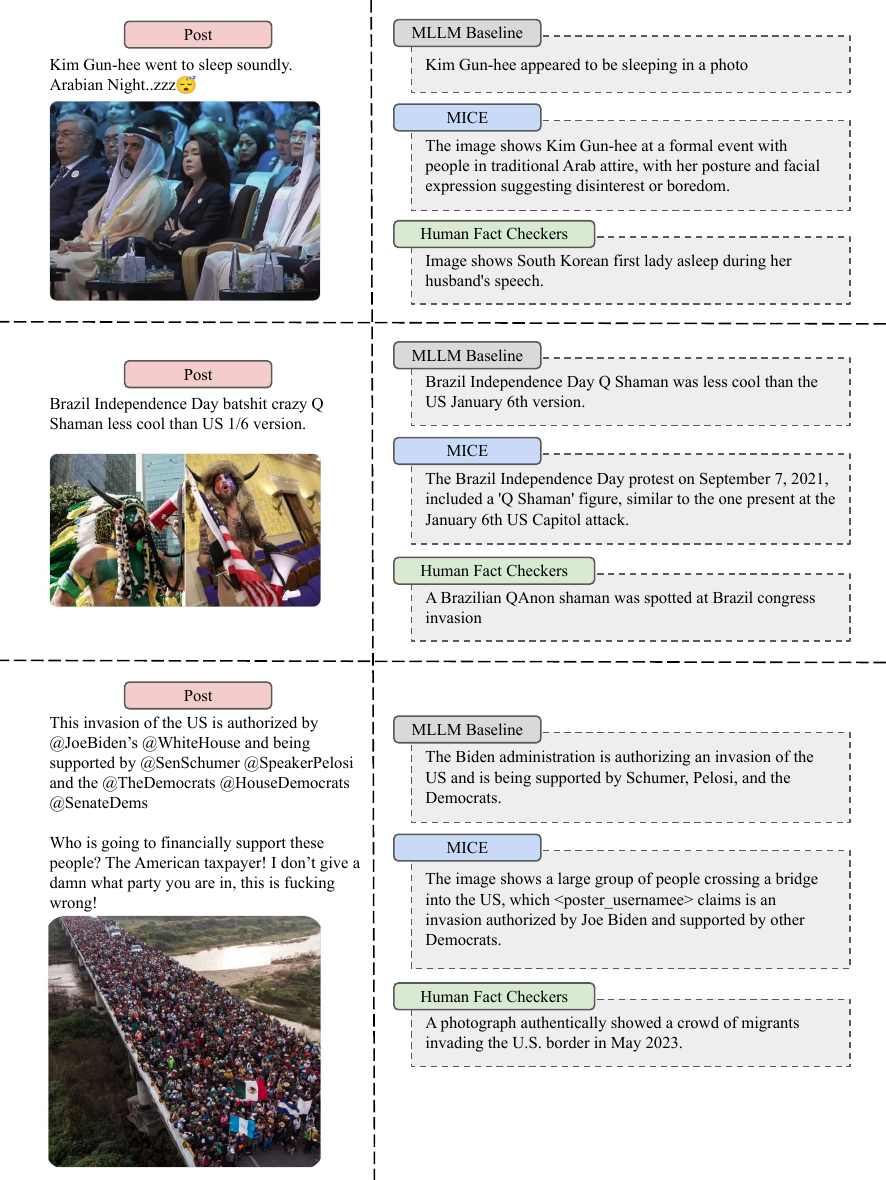}
    \caption{Examples of claim extraction from the MMCE dataset illustrating how baseline MLLMs (with in-context learning) underperform on intent-heavy multimodal posts. These cases highlight situations where structured reasoning (MICE) can help, but they are not representative of our dataset as a whole.}
    \label{fig:examples}
\end{figure}

The spread of misinformation on social media is increasingly multimodal, amplifying both its credibility and viral reach \citep{akhtar-etal-2023-multimodal}. Recent studies estimate that over one-third of debunked claims involve both text and images, highlighting the need for automated systems capable of reasoning across modalities \citep{zeng-etal-2024-multimodal, van-der-meer-etal-2025-hintsoftruth}. This shift has motivated interest in multimodal fact-checking \citep{10.1007/978-3-031-88720-8_68}, where prior work has primarily focused on claim detection \citep{cheema-etal-2022-mm} and verification \citep{10.1145/3539618.3591879, braun2025defame}. However, multimodal claim extraction remains largely unexplored, despite being a crucial step in the fact-checking process \cite{hassan2015detecting}. 

Extracting claims from multimodal content is challenging as systems must interpret informal language which often contains errors, integrate complementary or contradictory signals from images, and remove irrelevant context while preserving the factual core of a post \citep{wang-etal-2025-piecing}. Figure~\ref{fig:examples} shows examples in which the Gemini 2.0 Flash baseline fails to capture these nuances, as it tends to extract the claim literally as the text states or the image depicts, rather than inferring any underlying implication or context about the scenario. 

To address this gap, in this paper, we make the following contributions:
\begin{enumerate}
    \item We introduce MMCE (MultiModal Claim Extraction for fact checking), a new real-world benchmark dataset of 732 social media image-text posts paired with gold-standard claims, derived from fact-checking sources. 
    \item We introduce a scalable three-part evaluation framework for multimodal claim extraction, measuring semantic alignment with a manually annotated claim, faithfulness to the source, and decontextualization (i.e. whether a claim contains all necessary contextual information). 
    \item We provide a systematic study of MLLMs for claim extraction, showing that while they capture surface information, they often miss rhetorical intent and contextual cues.
    \item We introduce the Multimodal Intent-aware Claim Extraction framework (MICE), which is a practical framework that combines vision-based semantic analysis with MLLMs to perform intent- and context-aware claim extraction. MICE breaks down the claim extraction process, extracting visual information and modeling underlying intent and context before performing claim extraction, so that implicit claims can be surfaced rather than literal descriptions. Our analysis shows that MICE can act as a mitigation strategy for MLLMs when handling high-nuance, intent-critical claims.
\end{enumerate}

\section{MMCE (Multimodal Claim Extraction)}

\paragraph{Dataset}
We construct MMCE using post-claim pairs extracted from AVerImaTeC, a recently proposed dataset for real-world image-text claim verification based on data from fact-checking organizations \citep{cao2025averimatecdatasetautomaticverification}. The dataset focuses on out-of-context (OOC) image-text claims because recent studies had observed that context-manipulated claims are the most common type of media-based misinformation \citep{dufour2024ammebalargescalesurveydataset}. Additionally, extraction for OOC claims is the most different from traditional vision-language tasks, such as image captioning, where the image and the text align. The task deviates from the pre-training objectives of MLLMs, presenting a significant challenge.

Each post-claim pair in the dataset consists of the original text and image(s) from the social media post, as well as the extracted claim from the corresponding fact-checking article.
To do this, we first filter out data without links to the original social media posts, as well as data with claims that do not originate from social media sites. Next, we retrieve the source text directly from the linked social media posts. The final dataset contains 732 real-world post–claim pairs.\footnote{The final dataset includes 618 train / 114 dev pairs from AVerImaTeC; the original test split was not publicly available.}

For our experiment, we also curated a subset of 50 intent-critical claim examples that represent the failure modes that MICE is engineered to mitigate, to further analyze the utility of the intent-aware approach.

\paragraph{Evaluation}
In evaluating multimodal claim extraction, our goal is to go beyond simple similarity with a gold reference and instead capture multiple aspects of claim quality that matter for downstream verification in practice.
Thus, we draw on the insights from previous work, which demonstrate that LLMs can serve as effective evaluators, offering scalable, consistent, and context-sensitive judgments \citep{liu-etal-2023-g, fu-etal-2024-gptscore, muhamed2025ccrszeroshotllmasajudgeframework, es-etal-2024-ragas}. Moreover, existing work on the evaluation of text-only claim extraction goes beyond evaluating similarity, but also considers faithfulness to and coverage of the original content, as well as whether the claim contains all necessary contextual information \citep{metropolitansky-larson-2025-towards, ullrich2025claimextractionfactcheckingdata, deng-etal-2024-document}. These more comprehensive evaluation protocols provide a fuller picture of how well an extraction system supports downstream verification.

Motivated by these insights, we adopt an evaluation scheme consisting of both reference-based and reference-free metrics: \underline{(1) Reference-based} evaluation uses an LLM to judge the degree of semantic alignment between the generated claim and the gold reference claim; \underline{(2) Entailment} evaluation measures the extent to which the generated claim is faithful to the original social media post from which it was derived, while assuming that the post is true; \underline{(3) Decontextualization} evaluation determines whether the extracted claim is interpretable as a stand-alone factual statement, without requiring additional context. 

Our choice of model for evaluation is informed by previous studies, which have shown that Gemini models correlate relatively more with human assessments compared to other models \citep{akhtar2025ev2revaluatingevidenceretrieval, gu2025surveyllmasajudge}. As such, we use Gemini 2.5 Flash Lite as a judge.


\section{MICE (Multimodal Intent-aware Claim Extraction)}

\begin{figure}
    \centering
    \includegraphics[width=\linewidth]{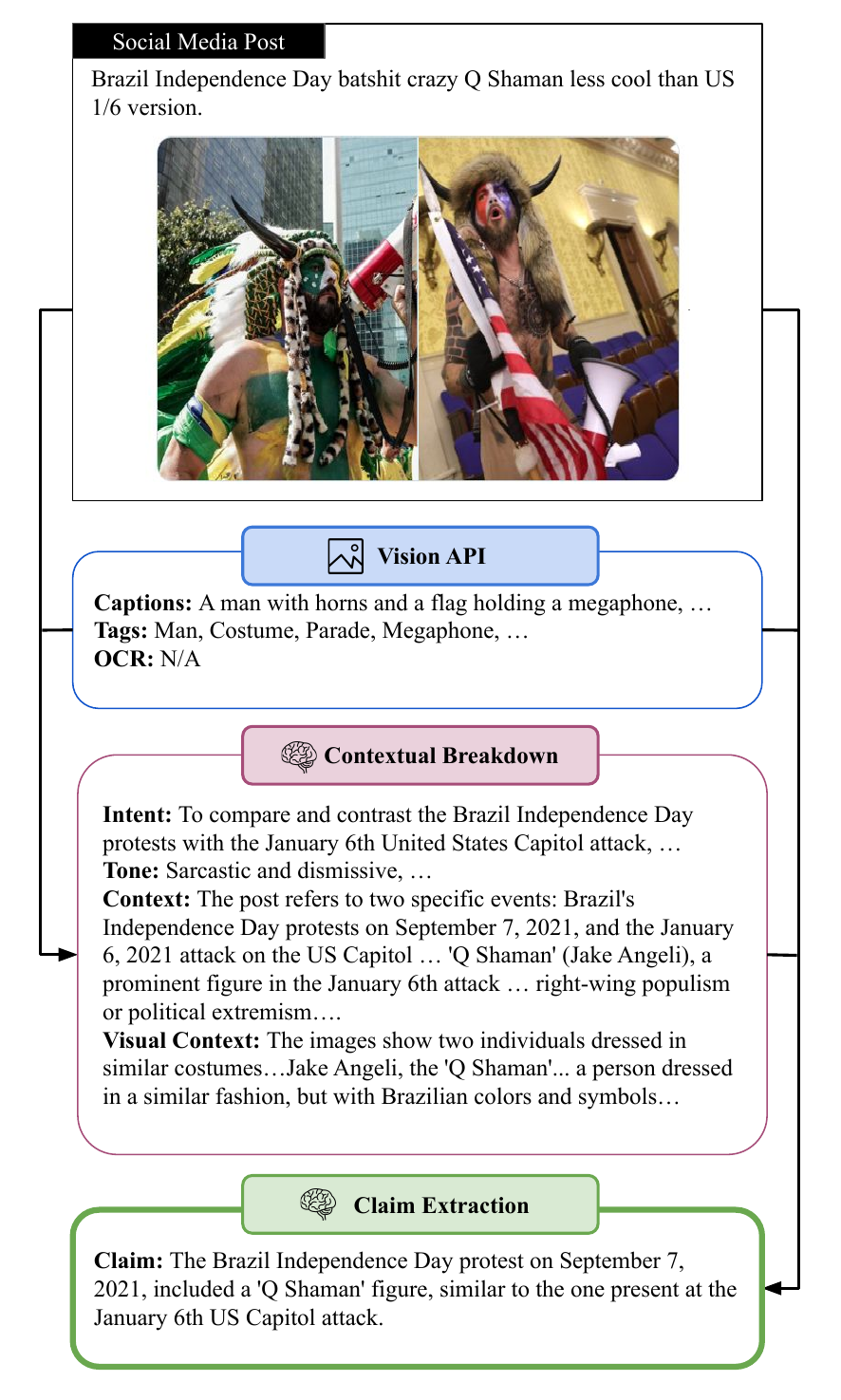}
    \caption{Overview of the MICE framework, which leverages visual understanding tools and MLLMs to reason across modalities.}
    \label{fig:architecture}
\end{figure}

We introduce MICE, a novel approach for extracting factual claims from social media posts. While baseline MLLMs can produce fluent claim outputs, our analysis shows that they often interpret text and image content too literally without inferring underlying intent and miss contextual or rhetorical cues. To address these shortcomings, MICE first extracts visual semantics from the image of the social media post to improve visual understanding, which is an approach supported by previous work showing that combining vision APIs with LLMs is effective for multimodal reasoning \citep{khademi-etal-2023-mm}. 
Next, MICE employs a contextual breakdown module to model intent, tone, and situational context, before integrating all signals to generate concise and accurate factual claims. The overall architecture 
is illustrated in Figure~\ref{fig:architecture}.

\paragraph{Vision API} Given an image-text social media post, we first apply the off-the-shelf Azure Vision API
to obtain structured descriptions of visual content. Specifically, we extract \underline{(1) Dense captions}: natural language descriptions of salient objects, attributes, and relations, \underline{(2) OCR text}: textual elements embedded in the image, and \underline{(3) Tags}: high-level labels (e.g, protest, flooding).

\paragraph{Contextual Breakdown} To capture intent and rhetorical context often present in social media posts, we prompt an MLLM for attributes which align with previous NLP-based frameworks for deception analysis, which models intention, tone, and contextual factors as core dimensions of deceptive content \citep{rani-etal-2025-sepsis}. To strike a balance between granular taxonomy and to avoid over-constraining the model outputs, we categorize the contextual breakdown into four crucial attributes: \underline{(1) Intention} captures what the author tries to achieve with the post (e.g. joking, criticizing), which helps distinguish factual claims from rhetorical or performative speech, \underline{(2) Tone} reflects the emotional or stylistic delivery (e.g. sarcastic, ironic, outraged), which signals whether a claim should be interpreted literally or as exaggeration, \underline{(3) Context} accounts for the surrounding situational and cultural cues (e.g. ongoing events/conflicts, pop culture), which provide the background necessary to construct a sound claim, and \underline{(4) Image Context} identifies specific visual elements (people, objects) shown in accompanying images, enabling the model to ground its understanding in both textual and visual information.

\paragraph{Claim Extraction} Finally, the claim extractor MLLM receives the post text and image(s), the visual representation, and the contextual information. It is prompted to produce factual claims with In-Context Learning (ICL), in which similar demonstrations are retrieved from the training data using BM25 \citep{10.1561/1500000019} under a leave-one-out strategy to prevent data leakage.

\section{Experiments}

\paragraph{Methods}
We evaluate a range of methods to establish performance across a diverse set of state-of-the-art MLLMs: Google Gemini 2.0 Flash \citep{gemini2024}, Qwen2.5 VL 32B Instruct \citep{bai2025qwen25vltechnicalreport}, and OpenAI GPT 4o Mini \citep{openai2024gpt4ocard}. We compare four methods: (1) MLLMs with text input only to evaluate the significance of visual information, (2) MLLMs with image-text inputs, (3) MLLMs with image-text inputs and ICL examples selected using BM25, and (4) the intent-aware framework, MICE. 

\paragraph{Main results}
\begin{table*}
\centering
\resizebox{\textwidth}{!}{%
    \begin{tabular}{lcccccc}  
    \hline
    \textbf{Method} & \textbf{Model (used in the method)} & \textbf{Reference-Based (1--4) ($\uparrow$)} & \multicolumn{2}{c}{\textbf{Entailment (\%)}} & \multicolumn{2}{c}{\textbf{Decontextualization (\%)}} \\
    \cline{4-5} \cline{6-7}
    & & & \textbf{Strict ($\uparrow$)} & \textbf{Lenient ($\uparrow$)} & \textbf{Strict ($\uparrow$)} & \textbf{Lenient ($\uparrow$)} \\
    \hline
    \rowcolor[gray]{0.65}  
    \hline
    \multicolumn{7}{c}{\textit{\textbf{MMCE dataset}}} \\    
    \hline
    \rowcolor[gray]{0.85}    
    \hline
    \multicolumn{7}{c}{\textit{LLM-based Evaluation}} \\    
    \hline
    \multirow{3}{*}{MLLM (text input only)} & Gemini 2.0 Flash & 2.80 & 80.0 & 85.9 & 96.9 & 99.7 \\
        & Qwen2.5 VL 32B Instruct & 2.85 & 77.7 & 86.3 & 96.4 & 99.6 \\
        & GPT 4o Mini & 2.83 & \textbf{80.6} & \textbf{88.0} & 97.4 & 99.7 \\
    \hline
    \multirow{3}{*}{MLLM} & Gemini 2.0 Flash & 3.11 & 75.5 & 85.4 & 97.9 & 99.9 \\
        & Qwen2.5 VL 32B Instruct & 3.14 & 77.3 & 86.6 & 98.5 & \textbf{100.0} \\
        & GPT 4o Mini & 3.15 & 74.4 & 85.2 & 97.8 & 99.9 \\
    \hline
    \multirow{3}{*}{MLLM with ICL} & Gemini 2.0 Flash & 3.21 & 70.5 & 82.6 & 98.4 & \textbf{100.0} \\
        & Qwen2.5 VL 32B Instruct & 3.24 & 71.8 & 83.6 & 98.4 & \textbf{100.0} \\
        & GPT 4o Mini & 3.22 & 69.9 & 82.4 & 98.1 & \textbf{100.0} \\
    \hline
    \multirow{3}{*}{MICE} & Gemini 2.0 Flash & \textbf{3.25} & 54.9 & 74.6 & \textbf{98.8} & \textbf{100.0} \\
        & Qwen2.5 VL 32B Instruct & 3.15 & 52.7 & 78.9 & 98.1 & \textbf{100.0} \\
        & GPT 4o Mini & 3.13 & 65.8 & 83.9 & 98.5 & \textbf{100.0} \\
    \hline
    \rowcolor[gray]{0.65}  
    \hline
    \multicolumn{7}{c}{\textit{\textbf{Intent-Critical Subset of MMCE}}} \\    
    \hline
    \rowcolor[gray]{0.85}    
    \hline
    \multicolumn{7}{c}{\textit{Human Evaluation}} \\    
    \hline
    MLLM with ICL & Gemini 2.0 Flash & 2.60 & 76.0 & 96.0 & 83.0 & \textbf{99.0} \\
    MICE & Gemini 2.0 Flash & \textbf{3.31} & \textbf{81.0} & \textbf{97.0} & \textbf{92.0} & \textbf{99.0} \\
    \hline
    \rowcolor[gray]{0.85}
    \hline
    \multicolumn{7}{c}{\textit{LLM-based Evaluation}} \\    
    \hline
    MLLM with ICL & Gemini 2.0 Flash & 2.08 & \textbf{68.0} & \textbf{82.0} & 94.0 & \textbf{100.0} \\
    MICE & Gemini 2.0 Flash & \textbf{3.56} & 48.0 & 68.0 & \textbf{100.0} & \textbf{100.0} \\
    \hline
    \end{tabular}
}
\caption{Experimental results in the full MMCE dataset, as well as the intent-critical subset of it. The bold values represent the best performing method for each dataset, and for each evaluation method (human and LLM-based). Scores in the \textit{Reference-Based} column are on a 1--4 scale (1 = lowest, 4 = highest). \textit{Entailment} and \textit{Decontextualization} are a 3-class categorical value, and is shown as strict (\% fully entailed / fully decontextualized) and lenient (\% fully or partially entailed / partially decontextualized).}
\label{tab:main_results}
\end{table*}

Table~\ref{tab:main_results} summarizes the results across the three evaluation metrics. In reference-based scoring, we find that MLLMs with image-text inputs consistently outperform instances with text-only inputs, underscoring the importance of incorporating visual signals for claim extraction in noisy social media posts. The further gains from MLLM with in-context learning (ICL) suggest that multimodal intent is best captured when models are guided by examples. 
In the subset of intent-critical cases, both human and LLM-based evaluations agree that the MICE framework improves reference-based scores, though its utility is not reflected across the entire dataset because of other challenges. E.g, when interpreting an image about the packaging of COVID-related medication, MLLMs and MICE fail to spot the insinuation that COVID is the same as influenza, which is indicative of the  challenges  MMCE poses to MLLMs more broadly
(see error analysis in Appendix~\ref{app:error_analysis}).

MICE also improves decontextualization scores for Gemini 2.0 flash and GPT 4o Mini, producing claims that are more stand-alone and interpretable without the original post. This is crucial for downstream fact-checking pipelines, where claims are often checked in isolation.
Our experiments also reveal that improvements in reference alignment and decontextualization introduce a trade-off with entailment. This is likely because explicitly modeling intent and context encourages the model to abstract away from the literal text and image. 

\paragraph{Human Alignment with Automated Evaluations}

To validate the LLM-based evaluation metrics and verify the utility of the MICE approach in handling intent-critical claims, we had 4 expert annotators independently score the claims extracted from the intent-critical subset (details in Appendix~\ref{app:annotation_guidelines}), and calculated the agreement between the human evaluators and the LLM scorer. 
The agreement is measured using Krippendorff's $\alpha$ \citep{krippendorff2013content} and Spearman's $\rho$ \citep{10.1093/ije/dyq191}. Results (Appendix~\ref{tab:human_alignment}) show that reference-based and decontextualization alignment achieved moderate correlation with humans ($\alpha=0.59$ and $\alpha=0.54$ respectively), but entailment score achieved a lower correlation ($\alpha=0.07$). 
This suggests that judging entailment involves a deeper inferential reasoning of the post compared to the other metrics, leading to a larger divergence from human judgment.

\paragraph{Experiment on Temporal Leakage}

To further examine the extent to which MLLMs rely on pre-training knowledge when generating claims, we compared Gemini 2.0 flash (training cutoff: June 2024) \citep{gemini2024} with Gemini 2.5 flash (training cutoff: January 2025) \citep{comanici2025gemini25pushingfrontier}. 
Experiments on 50 randomly sampled image–text claims posted between July and December 2024 (Appendix~\ref{app:temporal_leakage}) showed that Gemini 2.5 flash did not have a significant improvement in scoring metrics (Table~\ref{tab:temporal_leakage}). On closer analysis of the improved claims, we also found no clear evidence that Gemini 2.5 flash relied on additional pre-training knowledge unavailable to Gemini 2.0 flash.

\section{Conclusion}
In this work, we introduce MMCE, the first dataset for multimodal claim extraction \footnote{During the reviewing process we became aware of concurrent work by \cite{geng2026m4fcmultimodalmultilingualmulticultural}  that also proposes a dataset for this task.} from social media, and established a three-part evaluation framework that measures semantic alignment, faithfulness, and decontextualization.
Our analysis demonstrates that while baseline MLLMs can extract literal claims, they often miss the rhetorical intent and contextual framing crucial for understanding social media content.
To address this gap, we propose MICE, an intent-aware framework that improves performance in these nuanced, intent-critical cases. Overall, this research contributes a new benchmark and an intent-aware framework, supporting the ongoing development of more effective automated tools for fact-checking. 

\section*{Acknowledgments}
This research was supported by the  Alan Turing Institute and DSO National Laboratories in Singapore Partnership (ref  DCfP2\textbackslash100063). Zhenyun Deng, Zifeng Ding and Andreas Vlachos were further supported by the
ERC grant AVeriTeC (GA 865958). Andreas Vlachos is also supported by the DARPA program SciFy.
Michael Schlichtkrull is supported by the Engineering and Physical Sciences Research Council (grant number EP/Y009800/1), through funding from Responsible AI UK (KP0016).

\section*{Limitations}

We acknowledge that claim extraction can inherently be a subjective task, whereby extracted claims with different semantic meaning can still be considered valid and check-worthy. Our rationale is to model the claim extraction process as closely as possible to that of professional fact checkers, which motivated the decision to curate a dataset from real-world fact-checking articles.

Moreover, the dataset we curated explicitly focuses on out-of-context images, where we deemed it the most urgent to focus our efforts on these types of claim. However, this excludes image-text claim types that could also benefit from the MICE framework, such as memes. In future work, we propose to extend the experiments to other types of image-text claims for diversity.

Lastly, due to resource constraints, we performed human evaluation on 100 generated claims. Although this sample size provided a representative assessment of the claims' quality, it limits a more comprehensive statistical analysis of the framework's performance variability across different claim domains and styles. A larger scale human evaluation would be beneficial for future iterations of this work to improve statistical reliability and reproducibility.

\section*{Ethical Considerations}

We rely on fact-checks from real-world fact-checkers to develop and evaluate our models. Nevertheless, as any dataset, it is possible that it contains biases which influenced the development of
our approach. Given the societal importance of fact-checking, we advise that any automated system is
employed with human oversight to ensure that the
fact-checkers fact-check appropriate claims.

\bibliography{anthology,custom}

@misc{geng2026m4fcmultimodalmultilingualmulticultural,
      title={M4FC: a Multimodal, Multilingual, Multicultural, Multitask Real-World Fact-Checking Dataset}, 
      author={Jiahui Geng and Jonathan Tonglet and Iryna Gurevych},
      year={2026},
      eprint={2510.23508},
      archivePrefix={arXiv},
      primaryClass={cs.CL},
      url={https://arxiv.org/abs/2510.23508}, 
}

@inproceedings{hassan2015detecting,
    author = {Hassan, Naeemul and Li, Chengkai and Tremayne, Mark},
    title = {Detecting Check-worthy Factual Claims in Presidential Debates},
    year = {2015},
    isbn = {9781450337946},
    publisher = {Association for Computing Machinery},
    address = {New York, NY, USA},
    url = {https://doi.org/10.1145/2806416.2806652},
    doi = {10.1145/2806416.2806652},
    booktitle = {Proceedings of the 24th ACM International on Conference on Information and Knowledge Management},
    pages = {1835–1838},
    numpages = {4},
    location = {Melbourne, Australia},
    series = {CIKM '15}
}

@inproceedings{akhtar-etal-2023-multimodal,
    title = "Multimodal Automated Fact-Checking: A Survey",
    author = "Akhtar, Mubashara  and
      Schlichtkrull, Michael  and
      Guo, Zhijiang  and
      Cocarascu, Oana  and
      Simperl, Elena  and
      Vlachos, Andreas",
    editor = "Bouamor, Houda  and
      Pino, Juan  and
      Bali, Kalika",
    booktitle = "Findings of the Association for Computational Linguistics: EMNLP 2023",
    month = dec,
    year = "2023",
    address = "Singapore",
    publisher = "Association for Computational Linguistics",
    url = "https://aclanthology.org/2023.findings-emnlp.361/",
    doi = "10.18653/v1/2023.findings-emnlp.361",
    pages = "5430--5448"
}

@misc{akhtar2025ev2revaluatingevidenceretrieval,
      title={Ev2R: Evaluating Evidence Retrieval in Automated Fact-Checking}, 
      author={Mubashara Akhtar and Michael Schlichtkrull and Andreas Vlachos},
      year={2025},
      eprint={2411.05375},
      archivePrefix={arXiv},
      primaryClass={cs.CL},
      url={https://arxiv.org/abs/2411.05375}, 
}

@inproceedings{10.1007/978-3-031-88720-8_68,
author = {Alam, Firoj and Stru\ss{}, Julia Maria and Chakraborty, Tanmoy and Dietze, Stefan and Hafid, Salim and Korre, Katerina and Muti, Arianna and Nakov, Preslav and Ruggeri, Federico and Schellhammer, Sebastian and Setty, Vinay and Sundriyal, Megha and Todorov, Konstantin and V., Venktesh},
title = {The CLEF-2025 CheckThat! Lab: Subjectivity, Fact-Checking, Claim Normalization, and \& Retrieval},
year = {2025},
isbn = {978-3-031-88719-2},
publisher = {Springer-Verlag},
address = {Berlin, Heidelberg},
url = {https://doi.org/10.1007/978-3-031-88720-8_68},
doi = {10.1007/978-3-031-88720-8_68},
booktitle = {Advances in Information Retrieval: 47th European Conference on Information Retrieval, ECIR 2025, Lucca, Italy, April 6–10, 2025, Proceedings, Part V},
pages = {467–478},
numpages = {12},
location = {Lucca, Italy}
}

@misc{bai2025qwen25vltechnicalreport,
      title={Qwen2.5-VL Technical Report}, 
      author={Shuai Bai and Keqin Chen and Xuejing Liu and Jialin Wang and Wenbin Ge and Sibo Song and Kai Dang and Peng Wang and Shijie Wang and Jun Tang and Humen Zhong and Yuanzhi Zhu and Mingkun Yang and Zhaohai Li and Jianqiang Wan and Pengfei Wang and Wei Ding and Zheren Fu and Yiheng Xu and Jiabo Ye and Xi Zhang and Tianbao Xie and Zesen Cheng and Hang Zhang and Zhibo Yang and Haiyang Xu and Junyang Lin},
      year={2025},
      eprint={2502.13923},
      archivePrefix={arXiv},
      primaryClass={cs.CV},
      url={https://arxiv.org/abs/2502.13923},
}

@inproceedings{
braun2025defame,
title={{DEFAME}: Dynamic Evidence-based {FA}ct-checking with Multimodal Experts},
author={Tobias Braun and Mark Rothermel and Marcus Rohrbach and Anna Rohrbach},
booktitle={Forty-second International Conference on Machine Learning},
year={2025},
url={https://openreview.net/forum?id=umT6rMf1Rm}
}

@misc{cao2025averimatecdatasetautomaticverification,
      title={AVerImaTeC: A Dataset for Automatic Verification of Image-Text Claims with Evidence from the Web}, 
      author={Rui Cao and Zifeng Ding and Zhijiang Guo and Michael Schlichtkrull and Andreas Vlachos},
      year={2025},
      eprint={2505.17978},
      archivePrefix={arXiv},
      primaryClass={cs.CL},
      url={https://arxiv.org/abs/2505.17978}, 
}

@misc{gemini2024,
    author       = {DeepMind},
    title        = {Gemini 2.0 Flash Model Card},
    howpublished = {\url{https://modelcards.withgoogle.com/assets/documents/gemini-2-flash.pdf}},
    month        = dec,
    year         = {2024},
}

@misc{comanici2025gemini25pushingfrontier,
      title={Gemini 2.5: Pushing the Frontier with Advanced Reasoning, Multimodality, Long Context, and Next Generation Agentic Capabilities}, 
      author={DeepMind},
      year={2025},
      eprint={2507.06261},
      archivePrefix={arXiv},
      primaryClass={cs.CL},
      url={https://arxiv.org/abs/2507.06261}, 
}

@inproceedings{deng-etal-2024-document,
    title = "Document-level Claim Extraction and Decontextualisation for Fact-Checking",
    author = "Deng, Zhenyun  and
      Schlichtkrull, Michael  and
      Vlachos, Andreas",
    editor = "Ku, Lun-Wei  and
      Martins, Andre  and
      Srikumar, Vivek",
    booktitle = "Proceedings of the 62nd Annual Meeting of the Association for Computational Linguistics (Volume 1: Long Papers)",
    month = aug,
    year = "2024",
    address = "Bangkok, Thailand",
    publisher = "Association for Computational Linguistics",
    url = "https://aclanthology.org/2024.acl-long.645/",
    doi = "10.18653/v1/2024.acl-long.645",
    pages = "11943--11954"
}

@inproceedings{fu-etal-2024-gptscore,
    title = "{GPTS}core: Evaluate as You Desire",
    author = "Fu, Jinlan  and
      Ng, See-Kiong  and
      Jiang, Zhengbao  and
      Liu, Pengfei",
    editor = "Duh, Kevin  and
      Gomez, Helena  and
      Bethard, Steven",
    booktitle = "Proceedings of the 2024 Conference of the North American Chapter of the Association for Computational Linguistics: Human Language Technologies (Volume 1: Long Papers)",
    month = jun,
    year = "2024",
    address = "Mexico City, Mexico",
    publisher = "Association for Computational Linguistics",
    url = "https://aclanthology.org/2024.naacl-long.365/",
    doi = "10.18653/v1/2024.naacl-long.365",
    pages = "6556--6576"
}

@inproceedings{khademi-etal-2023-mm,
    title = "{MM}-Reasoner: A Multi-Modal Knowledge-Aware Framework for Knowledge-Based Visual Question Answering",
    author = "Khademi, Mahmoud  and
      Yang, Ziyi  and
      Frujeri, Felipe  and
      Zhu, Chenguang",
    editor = "Bouamor, Houda  and
      Pino, Juan  and
      Bali, Kalika",
    booktitle = "Findings of the Association for Computational Linguistics: EMNLP 2023",
    month = dec,
    year = "2023",
    address = "Singapore",
    publisher = "Association for Computational Linguistics",
    url = "https://aclanthology.org/2023.findings-emnlp.437/",
    doi = "10.18653/v1/2023.findings-emnlp.437",
    pages = "6571--6581"
}

@inproceedings{liu-etal-2023-g,
    title = "{G}-Eval: {NLG} Evaluation using Gpt-4 with Better Human Alignment",
    author = "Liu, Yang  and
      Iter, Dan  and
      Xu, Yichong  and
      Wang, Shuohang  and
      Xu, Ruochen  and
      Zhu, Chenguang",
    editor = "Bouamor, Houda  and
      Pino, Juan  and
      Bali, Kalika",
    booktitle = "Proceedings of the 2023 Conference on Empirical Methods in Natural Language Processing",
    month = dec,
    year = "2023",
    address = "Singapore",
    publisher = "Association for Computational Linguistics",
    url = "https://aclanthology.org/2023.emnlp-main.153/",
    doi = "10.18653/v1/2023.emnlp-main.153",
    pages = "2511--2522"
}

@inproceedings{metropolitansky-larson-2025-towards,
    title = "Towards Effective Extraction and Evaluation of Factual Claims",
    author = "Metropolitansky, Dasha  and
      Larson, Jonathan",
    editor = "Che, Wanxiang  and
      Nabende, Joyce  and
      Shutova, Ekaterina  and
      Pilehvar, Mohammad Taher",
    booktitle = "Proceedings of the 63rd Annual Meeting of the Association for Computational Linguistics (Volume 1: Long Papers)",
    month = jul,
    year = "2025",
    address = "Vienna, Austria",
    publisher = "Association for Computational Linguistics",
    url = "https://aclanthology.org/2025.acl-long.348/",
    doi = "10.18653/v1/2025.acl-long.348",
    pages = "6996--7045",
    ISBN = "979-8-89176-251-0"
}

@misc{openai2024gpt4ocard,
      title={GPT-4o System Card}, 
      author={OpenAI},
      year={2024},
      eprint={2410.21276},
      archivePrefix={arXiv},
      primaryClass={cs.CL},
      url={https://arxiv.org/abs/2410.21276}, 
}

@misc{ullrich2025claimextractionfactcheckingdata,
      title={Claim Extraction for Fact-Checking: Data, Models, and Automated Metrics}, 
      author={Herbert Ullrich and Tomáš Mlynář and Jan Drchal},
      year={2025},
      eprint={2502.04955},
      archivePrefix={arXiv},
      primaryClass={cs.CL},
      url={https://arxiv.org/abs/2502.04955}, 
}

@inproceedings{van-der-meer-etal-2025-hintsoftruth,
    title = "{H}ints{O}f{T}ruth: A Multimodal Checkworthiness Detection Dataset with Real and Synthetic Claims",
    author = "Van Der Meer, Michiel  and
      Korshunov, Pavel  and
      Marcel, S{\'e}bastien  and
      Plas, Lonneke Van Der",
    editor = "Che, Wanxiang  and
      Nabende, Joyce  and
      Shutova, Ekaterina  and
      Pilehvar, Mohammad Taher",
    booktitle = "Proceedings of the 63rd Annual Meeting of the Association for Computational Linguistics (Volume 1: Long Papers)",
    month = jul,
    year = "2025",
    address = "Vienna, Austria",
    publisher = "Association for Computational Linguistics",
    url = "https://aclanthology.org/2025.acl-long.1510/",
    doi = "10.18653/v1/2025.acl-long.1510",
    pages = "31274--31291",
    ISBN = "979-8-89176-251-0"
}

@inproceedings{10.1145/3539618.3591879,
    author = {Yao, Barry Menglong and Shah, Aditya and Sun, Lichao and Cho, Jin-Hee and Huang, Lifu},
    title = {End-to-End Multimodal Fact-Checking and Explanation Generation: A Challenging Dataset and Models},
    year = {2023},
    isbn = {9781450394086},
    publisher = {Association for Computing Machinery},
    address = {New York, NY, USA},
    url = {https://doi.org/10.1145/3539618.3591879},
    doi = {10.1145/3539618.3591879},
    booktitle = {Proceedings of the 46th International ACM SIGIR Conference on Research and Development in Information Retrieval},
    pages = {2733–2743},
    numpages = {11},
    location = {Taipei, Taiwan},
    series = {SIGIR '23}
}

@inproceedings{zeng-etal-2024-multimodal,
    title = "Multimodal Misinformation Detection by Learning from Synthetic Data with Multimodal {LLM}s",
    author = "Zeng, Fengzhu  and
      Li, Wenqian  and
      Gao, Wei  and
      Pang, Yan",
    editor = "Al-Onaizan, Yaser  and
      Bansal, Mohit  and
      Chen, Yun-Nung",
    booktitle = "Findings of the Association for Computational Linguistics: EMNLP 2024",
    month = nov,
    year = "2024",
    address = "Miami, Florida, USA",
    publisher = "Association for Computational Linguistics",
    url = "https://aclanthology.org/2024.findings-emnlp.613/",
    doi = "10.18653/v1/2024.findings-emnlp.613",
    pages = "10467--10484"
}

@misc{muhamed2025ccrszeroshotllmasajudgeframework,
      title={CCRS: A Zero-Shot LLM-as-a-Judge Framework for Comprehensive RAG Evaluation}, 
      author={Aashiq Muhamed},
      year={2025},
      eprint={2506.20128},
      archivePrefix={arXiv},
      primaryClass={cs.CL},
      url={https://arxiv.org/abs/2506.20128}, 
}

@inproceedings{wang-etal-2025-piecing,
    title = "Piecing It All Together: Verifying Multi-Hop Multimodal Claims",
    author = "Wang, Haoran  and
      Rangapur, Aman  and
      Xu, Xiongxiao  and
      Liang, Yueqing  and
      Gharwi, Haroon  and
      Yang, Carl  and
      Shu, Kai",
    editor = "Rambow, Owen  and
      Wanner, Leo  and
      Apidianaki, Marianna  and
      Al-Khalifa, Hend  and
      Eugenio, Barbara Di  and
      Schockaert, Steven",
    booktitle = "Proceedings of the 31st International Conference on Computational Linguistics",
    month = jan,
    year = "2025",
    address = "Abu Dhabi, UAE",
    publisher = "Association for Computational Linguistics",
    url = "https://aclanthology.org/2025.coling-main.498/",
    pages = "7453--7469"
}

@inproceedings{es-etal-2024-ragas,
    title = "{RAGA}s: Automated Evaluation of Retrieval Augmented Generation",
    author = "Es, Shahul  and
      James, Jithin  and
      Espinosa Anke, Luis  and
      Schockaert, Steven",
    editor = "Aletras, Nikolaos  and
      De Clercq, Orphee",
    booktitle = "Proceedings of the 18th Conference of the European Chapter of the Association for Computational Linguistics: System Demonstrations",
    month = mar,
    year = "2024",
    address = "St. Julians, Malta",
    publisher = "Association for Computational Linguistics",
    url = "https://aclanthology.org/2024.eacl-demo.16/",
    doi = "10.18653/v1/2024.eacl-demo.16",
    pages = "150--158"
}

@article{10.1561/1500000019,
    author = {Robertson, Stephen and Zaragoza, Hugo},
    title = {The Probabilistic Relevance Framework: BM25 and Beyond},
    year = {2009},
    issue_date = {April 2009},
    publisher = {Now Publishers Inc.},
    address = {Hanover, MA, USA},
    volume = {3},
    number = {4},
    issn = {1554-0669},
    url = {https://doi.org/10.1561/1500000019},
    doi = {10.1561/1500000019},
    journal = {Found. Trends Inf. Retr.},
    month = apr,
    pages = {333–389},
    numpages = {57}
}

@misc{gu2025surveyllmasajudge,
    title={A Survey on LLM-as-a-Judge}, 
    author={Jiawei Gu and Xuhui Jiang and Zhichao Shi and Hexiang Tan and Xuehao Zhai and Chengjin Xu and Wei Li and Yinghan Shen and Shengjie Ma and Honghao Liu and Saizhuo Wang and Kun Zhang and Yuanzhuo Wang and Wen Gao and Lionel Ni and Jian Guo},
    year={2025},
    eprint={2411.15594},
    archivePrefix={arXiv},
    primaryClass={cs.CL},
    url={https://arxiv.org/abs/2411.15594}, 
}

@book{krippendorff2013content,
  title={Content Analysis: An Introduction to Its Methodology},
  author={Krippendorff, K.},
  isbn={9781412983150},
  lccn={2011048278},
  url={https://books.google.com.sg/books?id=s_yqFXnGgjQC},
  year={2013},
  publisher={SAGE Publications}
}

@article{10.1093/ije/dyq191,
    author = {Spearman, C},
    title = {The proof and measurement of association between two things},
    journal = {International Journal of Epidemiology},
    volume = {39},
    number = {5},
    pages = {1137-1150},
    year = {2010},
    month = {10},
    issn = {0300-5771},
    doi = {10.1093/ije/dyq191},
    url = {https://doi.org/10.1093/ije/dyq191},
    eprint = {https://academic.oup.com/ije/article-pdf/39/5/1137/18481215/dyq191.pdf},
}

@misc{dufour2024ammebalargescalesurveydataset,
      title={AMMeBa: A Large-Scale Survey and Dataset of Media-Based Misinformation In-The-Wild}, 
      author={Nicholas Dufour and Arkanath Pathak and Pouya Samangouei and Nikki Hariri and Shashi Deshetti and Andrew Dudfield and Christopher Guess and Pablo Hernández Escayola and Bobby Tran and Mevan Babakar and Christoph Bregler},
      year={2024},
      eprint={2405.11697},
      archivePrefix={arXiv},
      primaryClass={cs.CY},
      url={https://arxiv.org/abs/2405.11697}, 
}

@inproceedings{rani-etal-2025-sepsis,
    title = "{SEPSIS}: {I} Can Catch Your Lies {--} A New Paradigm for Deception Detection",
    author = "Rani, Anku  and
      Dalal, Dwip  and
      Gautam, Shreya  and
      Gupta, Pankaj  and
      Jain, Vinija  and
      Chadha, Aman  and
      Sheth, Amit  and
      Das, Amitava",
    editor = "Zhao, Jin  and
      Wang, Mingyang  and
      Liu, Zhu",
    booktitle = "Proceedings of the 63rd Annual Meeting of the Association for Computational Linguistics (Volume 4: Student Research Workshop)",
    month = jul,
    year = "2025",
    address = "Vienna, Austria",
    publisher = "Association for Computational Linguistics",
    url = "https://aclanthology.org/2025.acl-srw.7/",
    doi = "10.18653/v1/2025.acl-srw.7",
    pages = "97--128",
    ISBN = "979-8-89176-254-1"
}
\bibliographystyle{acl_natbib}

\clearpage
\appendix

\section{MMCE Dataset}
\label{sec:mmce}
\renewcommand{\thetable}{A\arabic{table}}
\setcounter{table}{0} 

The breakdown of the social media sites of the data from MMCE is shown in table~\ref{tab:social_media_sites}.

\begin{table}[h]
\centering
\begin{tabular}{lc}
\hline
\textbf{Social Media Site} & \textbf{Number of Claims} \\
\hline
X (Formally Twitter) & 360 \\
Facebook & 319 \\
Instagram & 32 \\
Reddit & 13 \\
Telegram & 2 \\
Weibo & 2 \\
Band (Naver) & 1 \\
Flickr & 1 \\
Truth Social & 1 \\
\hline
\end{tabular}
\caption{Breakdown of social media sites}
\label{tab:social_media_sites}
\end{table}

Although we converted the original post text to English for our experiments, we also retained the original language of the social media posts for future experiments. Table ~\ref{tab:langauges} shows the breakdown of the language of the original posts in MMCE.

\begin{table}[h]
\centering
\begin{tabular}{lc}
\hline
\textbf{Original Post Language} & \textbf{Number of Claims} \\
\hline
English & 551 \\
Hindi & 100 \\
Korean & 14 \\
Urdu & 10 \\
Chinese & 8 \\
Bengali & 7 \\
Thai & 6 \\
Sinhala & 6 \\
Khmer & 7 \\
Burmese & 6 \\
Filipino & 4 \\
Amharic & 4 \\
Tamil & 3 \\
Indonesian & 2 \\
Punjabi & 2 \\
Telugu & 2 \\
Kannada & 1 \\
French & 1 \\
Pashto & 1 \\
Swahili & 1 \\
\hline
\end{tabular}
\caption{Breakdown of the language of the original posts in MMCE. Some posts use more than one language. In those cases, we count it under all the languages used.}
\label{tab:langauges}
\end{table}

\section{Human Alignment Analysis}
\label{app:human_alignment}
\renewcommand{\thetable}{B\arabic{table}}
\setcounter{table}{0} 

Table~\ref{tab:human_alignment} reports detailed agreement statistics between LLM-based evaluations and human annotators, as well as human–human agreement.

\begin{table*}
\centering
\resizebox{\textwidth}{!}{%
    \begin{tabular}{lcccccc}  
    \hline
    \textbf{Method} & \textbf{Model} & \textbf{Reference-Based} & \textbf{Entailment} & \textbf{Decontextualization} \\
    \hline
    \multirow{3}{*}{LLM-Human} 
    & Krippendorff’s $\alpha$ & 0.59 & 0.07 & 0.54 \\
    & Spearman $\rho$ & 0.56 & 0.11 & 0.41 \\
    & Agreement (\%) & 46.5 & 50.0 & 89.5 \\
    \hline
    \multirow{3}{*}{Human-Human} 
    & Krippendorff’s $\alpha$ & 0.67 & 0.14 & 0.80 \\
    & Spearman $\rho$ & 0.70 & 0.30 & 0.81 \\
    & Agreement (\%) & 55.0 & 69.0 & 94.0 \\
    \hline
    \end{tabular}
}
\caption{Human–LLM and human–human alignment statistics across the three evaluation metrics used in the experiment.}
\label{tab:human_alignment}
\end{table*}

Due to resource constraints, one annotator labeled all 100 examples; the other three split the set (approximately 33–34 claims each).

\section{Temporal Leakage Analysis}
\label{app:temporal_leakage}
\renewcommand{\thetable}{C\arabic{table}}
\setcounter{table}{0} 

Data for the temporal leakage analysis was drawn from four fact-checking sources via the Google Fact Check API. The breakdown of the source URLs of the data used in the temporal leakage analysis is shown in table~\ref{tab:fact_checking_sites}.

\begin{table}[H]
\centering
\begin{tabular}{lc}
\hline
\textbf{Fact Checking Site} & \textbf{Number of Claims} \\
\hline
factcheck.afp.com & 30 \\
leadstories.com & 13 \\
politifact.com & 5 \\
fullfact.org & 2 \\
\hline
\end{tabular}
\caption{Breakdown of fact checking sites used in the temporal leakage experiments}
\label{tab:fact_checking_sites}
\end{table}

The results of the temporal leakage experiment are shown in table~\ref{tab:temporal_leakage}.
\begin{table*}[h]
\centering
\resizebox{\textwidth}{!}{%
    \begin{tabular}{lccccc}
    \hline
    \textbf{Model} & \textbf{Reference-Based (1--4) ($\uparrow$)} & \multicolumn{2}{c}{\textbf{Entailment (\%)}} & \multicolumn{2}{c}{\textbf{Decontextualization (\%)}} \\
    \cline{3-4} \cline{5-6}
    & & \textbf{Strict ($\uparrow$)} & \textbf{Lenient ($\uparrow$)} & \textbf{Strict ($\uparrow$)} & \textbf{Lenient ($\uparrow$)} \\
    \hline
    Gemini 2.0 Flash & 3.08 & \textbf{88.0} & \textbf{90.0} & \textbf{94.0} & \textbf{98.0} \\
    Gemini 2.5 Flash & \textbf{3.10} & 86.0 & 86.0 & 84.0 & 90.0 \\
    \hline
    \end{tabular}
}
\caption{Results of the temporal leakage experiment, on 50 post-claim pairs that surfaced after the training cut-off for Gemini 2.0 Flash, but before the training cut-off for Gemini 2.5 Flash. Scores in the \textit{Reference-Based} column are on a 1--4 scale (1 = lowest, 4 = highest). \textit{Entailment} and \textit{Decontextualization} are shown as strict (\% fully entailed / fully decontextualized) and lenient (\% fully or partially entailed / partially decontextualized).}
\label{tab:temporal_leakage}
\end{table*}

\section{Model Implementation and Resources}
\label{app:resources}
For few-shot settings, we set the number of shots to be 5 to strike a balance between providing sufficient demonstrations, and to prevent exceeding the context length limit.

For experiments with Qwen2.5 VL 32B Instruct, we faced a 30-image limit per input. In the handful of cases where the limit is exceeded (due to the images from the few shot demonstrations), we limit the number of few-shot inputs to fit the image constraints.

All experiments were conducted using API access to commercial and open-source multimodal LLMs via the OpenRouter\footnote{https://openrouter.ai/} platform. We report results for Gemini 2.0 Flash, Gemini 2.5 Flash, Qwen2.5 VL 32B Instruct, and GPT-4o Mini. Since these models are hosted services, we do not control the underlying hardware; however, we record the model versions and set temperature to zero to facilitate reproducibility.

\section{Licensing Information}
\label{app:license}
If accepted, we will publicly release and maintain the dataset and baseline code, which will be licensed under the CC BY-NC 4.0 license.

\section{Prompts}
\label{app:prompts}
Here we provide the prompts used in our experiments. This includes the prompts used in the LLM-based evaluation, as well as the prompts used in the baseline approaches and the MICE framework.

For the MICE framework, the prompt used in the claim extraction stage is a concatenation of the image-text input prompt (~\ref{app:image-text-CE-prompt}), the output of the Vision API, and the output of the contextual extraction (~\ref{app:context-prompt}).

\subsection{Prompt for reference-based evaluation}
\begin{lstlisting}[style=FlushLeftPrompt]
# MISSION
You will get two claims below, a generated claim and a reference claim. Your task is to perform a comprehensive similarity assessment between the generated claim and the reference claim, in the context of fact checking. Provide a similarity score from 1 to 4.

# CRITICAL INSTRUCTIONS:
- Be CONSISTENT in your scoring. Similar claims should always receive similar scores.
- Focus on the core implied meaning/content of the sentence, wording differences should be acceptable.
- Extra details that don't contradict the core claim is beneficial. Additional specific details (e.g. names, dates, locations, numbers) on the core factual assertion is a good thing. Do not penalize for verbosity and specificity.

# Scoring Guidelines:
- 1: Completely different, no overlap in themes, topics, or entities mentioned.
- 2: Minimal similarity in themes, topics, or entities mentioned, sentences have different meanings. The core factual statement to be fact checked is different.
- 3: Partial alignment in message conveyed, with significant differences that could potentially affect downstream fact checking.
- 4: Strong conceptual similarity with minor variations or near-identical meaning. 

# RESPONSE FORMAT
Return the response in the following JSON format:
```json
{
    "score": 3,
    "reasoning": "Brief explanation of why this score was given"
}
```

# INPUT
Generated Claim: <generated claim>
Reference Claim: <reference claim>
\end{lstlisting}

\subsection{Prompt to evaluate entailment}
\begin{lstlisting}[style=FlushLeftPrompt]
# MISSION
You will be given a social media post via the text and image of the post, we well as a candidate claim extracted from the post. Your task is to assess whether the claim is fully faithful to and entailed by the combined content of the image and text. This means that assuming the social media post is true, the extracted claim must also be true.

Do NOT check if the claim is true in reality, only whether it is faithful to the content of the image and text of the post.

Ignore whether the correct factual content had been extracted, focus on whether the extracted sentence is faithful (i.e. no hallucinations).

# EVALUATION CRITERIA
Classify the claim into one of three categories:
- **entailed**: The claim is fully aligned with the post content without any contradictions, hallucinations or unsupported additions.
- **partially_entailed**: The claim is partially aligned with the post content but contains minor variations or additional context not stated or implied in the post.
- **not_entailed**: The claim contains significant misaligned inferences, exaggerations beyond what's stated, major contradictions, hallucinations, or completely misaligns with the post content.

# INPUT
Generated Claim: <generated claim>
Text: <social media post text>
Image(s): <social media post image(s)>
\end{lstlisting}

\subsection{Prompt to evaluate decontextualization}
\begin{lstlisting}[style=FlushLeftPrompt]
# MISSION
You will be given a candidate claim that was extracted from a social media post. Your task is to assess whether this claim is understandable in isolation, without access to the original post or any external context. A decontextualized claim should be fully interpretable and self-contained to an average reader, who has no knowledge of the post.

# CRITICAL INSTRUCTIONS:
- Focus only on clarity and completeness of meaning. Do not check whether the claim is factually true or faithful to the post, only whether the claim can stand alone and be understood independently.

# Scoring Guidelines:
- **fully_decontextualized**: Understandable in isolation. The claim is completely self-contained, unambiguous, and requires no edits to be understood on its own. (Example: The mayor of NYC announced a new recycling program on June 1, 2024.)
- **partially_decontextualized**: The claim is mostly clear and contains some context, but has gaps, vague references or unresolved pronouns. The claim could benefit from some edits. (Example: Vaccination rates rose after that. -> could be rewrited to -> Vaccination in the UK rates rose after the 2023 campaign.)
- **not_decontextualized**: Not understandable in isolation. The claim cannot be interpreted on its own; key entities, referents, or context are missing. Major rewriting is needed. (Example: He did something yesterday.)

# INPUT
Generated Claim: <generated claim>
\end{lstlisting}

\subsection{Prompt for text-only input for claim extraction}
\begin{lstlisting}[style=FlushLeftPrompt]
# MISSION
You are an expert fact-checking analyst specializing in social media content verification. Your primary objective is to precisely extract and articulate the core factual claim(s) from the given text.

# CONTEXT ANALYSIS
Before extracting the claim, perform a comprehensive context analysis:
- Examine the full text carefully
- Consider the platform type and its typical communication style
- Identify potential implicit or explicit claims

# CLAIM EXTRACTION METHODOLOGY
1. Identify Potential Claims:
   - Look for definitive statements
   - Detect implied assertions
   - Recognize potentially misleading or exaggerated claims

2. Claim Criteria:
   - Clarity: Can the claim stand alone and be understood without the original context?
   - Specificity: Does the claim capture the most significant factual assertion?
   - Verifiability: Does the claim provide enough detail to enable fact-checking?

3. Claim Refinement Process:
   - Remove subjective language
   - Distill the core factual assertion
   - Ensure the claim is neutral and objective

# CLAIM SELECTION STRATEGY
- Always try to extract just one main claim first
- If the text contains one main factual assertion, extract only that claim
- If multiple statements can be combined into one coherent claim, do so

Multiple claims should only be used when:
- The text contains completely separate factual statements about different topics that cannot be combined
- Each claim is independently verifiable and fact-checkable
- Combining them would create a confusing or overly complex claim

# ADDITIONAL CONSIDERATIONS
- If multiple potential claims exist, first try to identify the most significant or impactful one
- If the claim is ambiguous, provide the most reasonable interpretation based on context
- Avoid introducing personal bias or speculation
- Always prioritize extracting a single, comprehensive claim over multiple separate claims

# RESPONSE FORMAT
Return the response in the following JSON format:
```json
{
    "claims": ["main claim"]
}

# INPUT
Extract the claim(s) from the following text: <social media post text>
\end{lstlisting}

\subsection{Prompt for image-text input for claim extraction}
\label{app:image-text-CE-prompt}
\begin{lstlisting}[style=FlushLeftPrompt]
# MISSION
You are an expert fact-checking analyst specializing in social media content verification. Your primary objective is to precisely extract and articulate the core factual claim(s) from the given text and accompanying image.

# CONTEXT ANALYSIS
Before extracting the claim, perform a comprehensive context analysis:
- Examine the full input text and image carefully. Consider how the image contributes to the messaging
- Consider the platform type and its typical communication style
- Identify potential implicit or explicit claims

# CLAIM EXTRACTION METHODOLOGY
1. Identify Potential Claims:
   - Look for definitive statements
   - Detect implied assertions
   - Recognize potentially misleading or exaggerated claims

2. Claim Criteria:
   - Clarity: Can the claim stand alone and be understood without the original context?
   - Specificity: Does the claim capture the most significant factual assertion?
   - Verifiability: Does the claim provide enough detail to enable fact-checking?

3. Claim Refinement Process:
   - Remove subjective language
   - Distill the core factual assertion
   - Ensure the claim is neutral and objective
   - Consider whether the image alters, reinforces, or creates the perceived claim

# CLAIM SELECTION STRATEGY
- Always try to extract just one main claim first
- If the text contains one main factual assertion, extract only that claim
- If multiple statements can be combined into one coherent claim, do so

Multiple claims should only be used when:
- The text contains completely separate factual statements about different topics that cannot be combined
- Each claim is independently verifiable and fact-checkable
- Combining them would create a confusing or overly complex claim
- The image introduces additional factual assertions that cannot be combined with the text claims


# ADDITIONAL CONSIDERATIONS
- If multiple potential claims exist, first try to identify the most significant or impactful one
- If the claim is ambiguous, provide the most reasonable interpretation based on context
- Avoid introducing personal bias or speculation
- Always prioritize extracting a single, comprehensive claim over multiple separate claims
- Consider whether the image alters, reinforces, or creates the perceived claim

# RESPONSE FORMAT
Return the response in the following JSON format:
```json
{
    "claims": ["main claim"]
}
```

# INPUT
Extract the claim(s) from the following text: <social media post text>
\end{lstlisting}

\subsection{Prompt to extract contextual insights}
\label{app:context-prompt}
\begin{lstlisting}[style=FlushLeftPrompt]
# MISSION
Analyze this social media post and provide contextual insights to help identify the main factual claim.

# CONTEXTUAL ANALYSIS
Focus on these key insights:
1. INTENT: What's the main purpose of the post? (inform/persuade/entertain/satire/etc.)
2. TONE: What's the emotional tone of the post? (serious/humorous/sarcastic/anger/etc.)
3. CONTEXT: What real-world events/issues does this relate to? Include specific details like dates, locations, people, organizations.
4. VISUAL_CONTEXT: What specific people, objects, locations, or events are shown in the image that provide context for the claim? 

# RESPONSE FORMAT
Return your analysis as a JSON object with the following structure:
```json
{
    "intent": "description of the poster's main purpose",
    "tone": "description of the emotional tone",
    "context": "brief context with specific details about real-world events/issues",
    "visual_context": "description of what's shown in the image that provides context"
}
```
\end{lstlisting}

\section{Annotation Guidelines}
\label{app:annotation_guidelines}
The annotators involved in the human evaluation are also co-authors but were unaware of which system generated which claim. Here, we provide the annotation guidelines provided to those annotators, alongside an excel sheet containing the intent-critical subset of the 100 post-claim pairs, as well as the images relevant to the posts.

\subsection{Instructions given to the annotators}
\begin{lstlisting}[style=FlushLeftPrompt]
There are 100 generated claims in the excel file, and for each claim, the original social media post (text+image(s)) and the reference claim, are provided. Fill in the green columns (E and F): reference-based score and entailment score.

Here are the guidelines for the scoring:

Reference-based 
[Look at columns C and D only]
Given 2 claims, a generated claim and a reference claim, perform a comprehensive similarity assessment between the generated claim and the reference claim. Provide a similarity score from 1 to 4.

Focus on the core meaning and factual content, not minor wording differences. Ignore extra  details that don't change the core factual claim, no need to penalize for that.

Scoring Guideline (1-4 scale)
1: Completely different, no overlap in themes, topics, or entities mentioned.
2: Minimal similarity in themes, topics, or entities mentioned, sentences have different meanings. The core factual statement to be fact checked is different.
3: Partial alignment in message conveyed, with significant differences that could potentially affect downstream fact checking.
4: Strong conceptual similarity with minor variations or near-identical meaning. 

Entailment
[Look at columns A, B and D only]
Given a social media post (text and image(s)), as well as a claim extracted from the post, assess whether the claim is fully faithful to and entailed by the combined content of the image and text. This means that assuming the social media post is true, the extracted claim must also be true.

Ignore whether the correct factual content had been extracted, focus on whether the extracted sentence is faithful to the post (i.e. no hallucinations).

Scoring Guideline (3-class classification)
1. entailed: The claim is fully aligned with the post content without any contradictions, hallucinations or unsupported additions.
2. partially_entailed: The claim is partially aligned with the post content but contains minor variations or additional context not stated or implied in the post.
3. not_entailed: The claim contains significant misaligned inferences, exaggerations beyond what's stated, major contradictions, hallucinations, or completely misaligns with the post content.

Decontextualization
[Look at column D only]
You will be given a candidate claim that was extracted from a social media post. Your task is to assess whether this claim is understandable in isolation, without access to the original post or any external context. A decontextualized claim should be fully interpretable and self-contained to an average reader, who has no knowledge of the post.
Scoring Guideline (3-class classification)
1. fully_decontextualized: Understandable in isolation. The claim is completely self-contained, unambiguous, and requires no edits to be understood on its own. (Example: The mayor of NYC announced a new recycling program on June 1, 2024.)
2. partially_decontextualized: The claim is mostly clear and contains some context, but has gaps, vague references or unresolved pronouns. The claim could benefit from some edits. (Example: Vaccination rates rose after that. -> could be rewrited to -> Vaccination in the UK rates rose after the 2023 campaign.)
3. not_decontextualized: Not understandable in isolation. The claim cannot be interpreted on its own; key entities, referents, or context are missing. Major rewriting is needed. (Example: He did something yesterday.) 
\end{lstlisting}

\section{Error Analysis}
\label{app:error_analysis}
\renewcommand{\thetable}{H\arabic{table}}
\setcounter{table}{0} 

To surface the key challenges faced in image-text social media claim extraction, we select erroneous samples (scored 1 or 2 for reference-based scoring) for analysis, shown in Table~\ref{tab:error_analysis_part1}. We also select instances where the baseline MLLM with ICL outperforms MICE in reference-based scoring, shown in Table~\ref{tab:mice_failures}, in order to highlight some potential drawbacks of MICE. Note that in cases where the system outputs more than one claim, we select the highest scoring claim based on the reference-based evaluation.
\begin{table*}[t]
\centering
\label{tab:smallerfont}
\footnotesize
\begin{tabularx}{\textwidth}{XXX}
    \hline
    \rowcolor[gray]{0.7}  
    \hline
    \multicolumn{3}{c}{\textit{\textbf{Failure Cases for both baseline MLLMs and MICE}}} \\    
    \hline
    \textbf{Social Media Post} & \textbf{Claims} & \textbf{Analysis} \\
    \hline
        India's clothing minister - Prime Minister of Italy. Very cultured. \includegraphics[width=5cm]{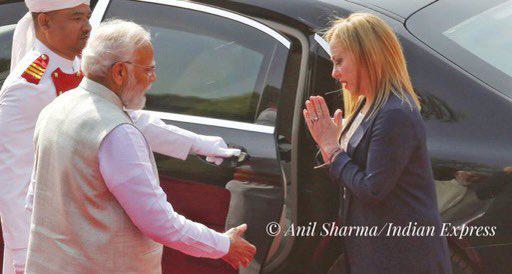}
        & \textcolor{ForestGreen}{Reference:} Italy's Prime Minister Giorgia Meloni refuses to shake hands with PM Narendra Modi. \newline\newline
        \textcolor{Mulberry}{MLLM+ICL:} Image shows Narendra Modi and the Prime Minister of Italy. [2] \newline\newline
        \textcolor{MidnightBlue}{MICE:} The image shows Narendra Modi opening a car door for Italian Prime Minister Giorgia Meloni. [1]
        & The LLM-based approaches fail to capture the nuance of the gestures and context in the image. The phrase “Very cultured” adds a subjective commentary on behavior rather than merely describing the scene. Though MICE attempts to interpret the actions of the people in the image, it overlooks intended meaning of the post. This reflects a broader limitation in MLLMs' ability to integrate non-literal cues from language with the contextual visual semantics of human gestures.\\
    \hline
        Krabi people are enlightened!!! Move Forward Party is no longer a party that will overthrow the monarchy. \#MoveForward \#themalaengtad \includegraphics[width=5cm]{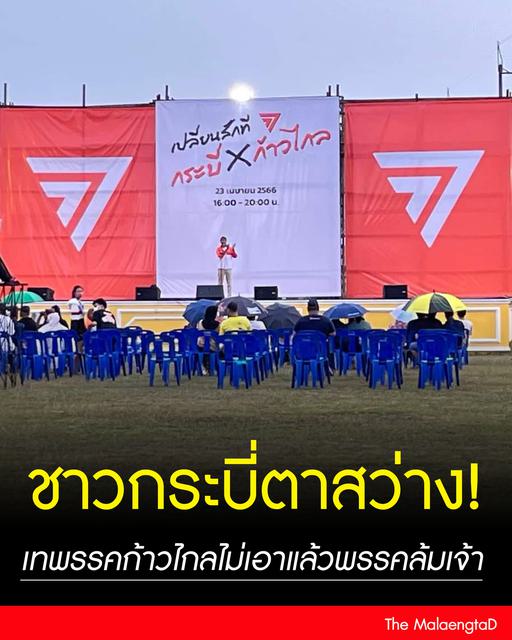} \newline \textit{The text overlay in the image is: "Krabi people's eyes are open! If you don't take the party forward, the party will fall."}
        & \textcolor{ForestGreen}{Reference:} The Move Forward Party's political rally in Southern Krabi province, Thailand had a very small crowd attendance. \newline\newline
        \textcolor{Mulberry}{MLLM+ICL:} The Move Forward Party is no longer a party that will overthrow the monarchy. [1] \newline\newline
        \textcolor{MidnightBlue}{MICE:} On April 23, 2023, the Move Forward Party held an event in Krabi, and the party is no longer viewed as wanting to overthrow the monarchy. [2]
        & The LLM-based models misinterpret the underlying sarcasm and political sentiment embedded in the post. The Thai text, combined with the image of a sparse crowd, is meant to mock the Move Forward Party's rally attendance. However, both MICE and MLLM+ICL treat the post as a straightforward political statement, failing to synthesize the ironic signals between text and visuals to understand the true rhetorical intent. \\
    \hline
        Packaging update \includegraphics[height=1em]{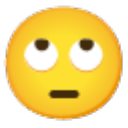} \includegraphics[width=5cm]{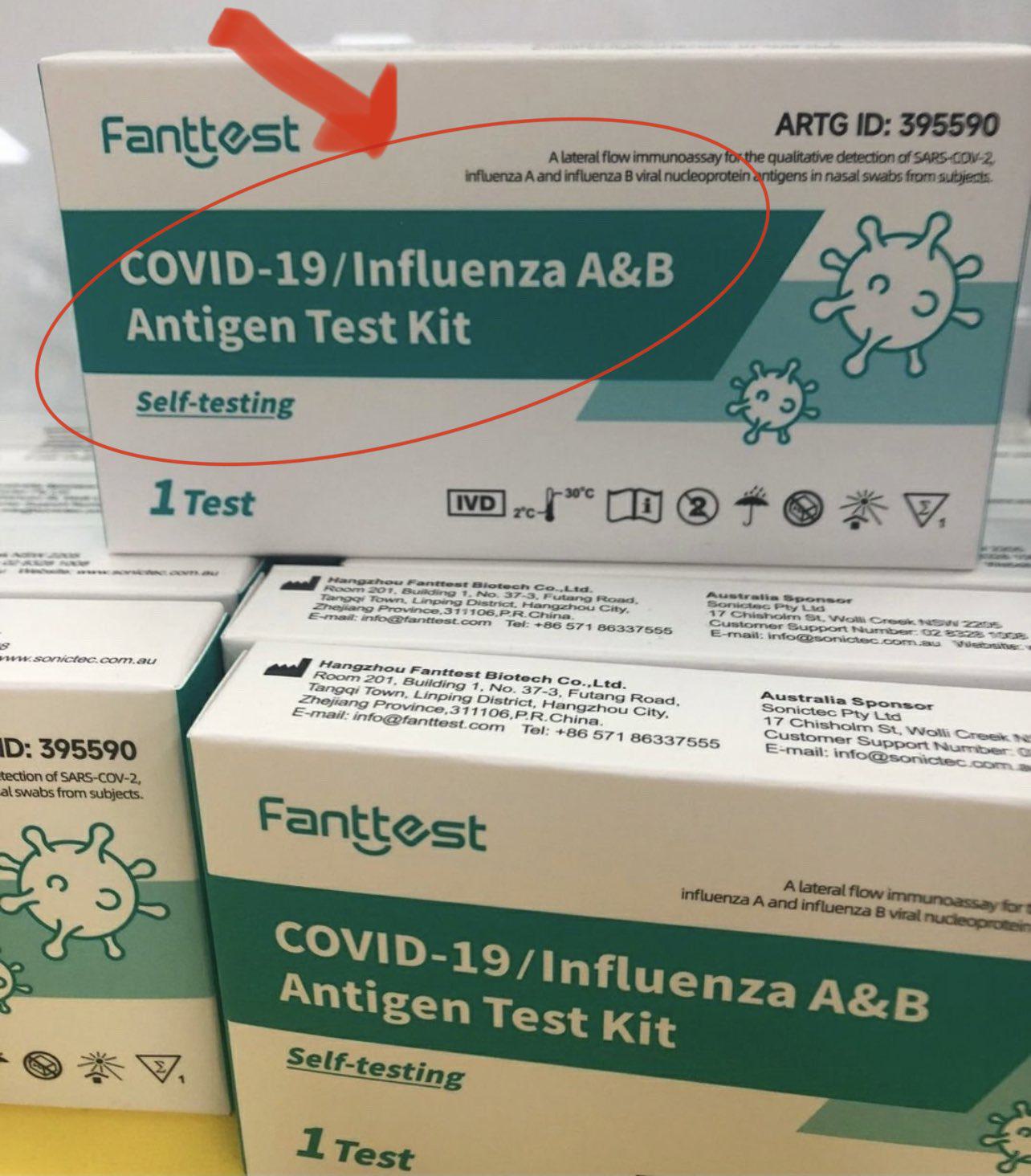}
        & \textcolor{ForestGreen}{Reference:} COVID19 is the same as influenza. \newline\newline
        \textcolor{Mulberry}{MLLM+ICL:} Fanttest sells a COVID-19/Influenza A\&B Antigen Test Kit. [1] \newline\newline
        \textcolor{MidnightBlue}{MICE:} The packaging for the Fanttest COVID-19/Influenza A\&B Antigen Test Kit has been updated. [1] 
        & The LLM-based approaches fail to identify the implicit misinformation within the post. Both MICE and MLLM+ICL interpret the post literally without recognizing the underlying insinuation that conflates two distinct diseases. The emoji also serves as a rhetorical device signaling skepticism and insinuation, which was not picked up by the systems. Consequently, their outputs remain factually descriptive but semantically shallow, demonstrating a persistent limitation in detecting implicit misinformation. \\
    \hline
\end{tabularx}
\caption{Analysis on instances where both baseline MLLMs and the MICE framework yields poor results, using Gemini 2.0 Flash as the underlying model. The numbers within the square brackets represent the LLM-based reference based score \textbf{(Part 1)}.}
\label{tab:error_analysis_part1}
\end{table*}

\begin{table*}[t]
\centering
\label{tab:smallerfont}
\footnotesize
\begin{tabularx}{\textwidth}{XXX}
    \hline
    \rowcolor[gray]{0.7}  
    \hline
    \multicolumn{3}{c}{\textit{\textbf{Failure Cases for both baseline MLLMs and MICE}}} \\    
    \hline
    \textbf{Social Media Post} & \textbf{Claims} & \textbf{Analysis} \\
    \hline
        The name of this Chinese firm is PMC Projects and its owner is 'Chang Chien-ting', better known as Morris Chang. Surprisingly, Morris Chang is the son of Chinese citizen Chang Chung-ling, who has been a director of Adani's companies and is a business partner of Adani's brother Vinod Adani. \includegraphics[width=5cm]{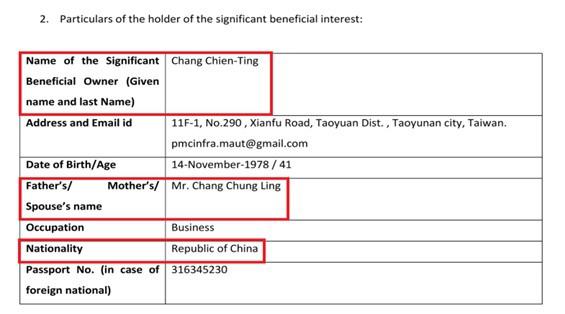} \includegraphics[width=5cm]{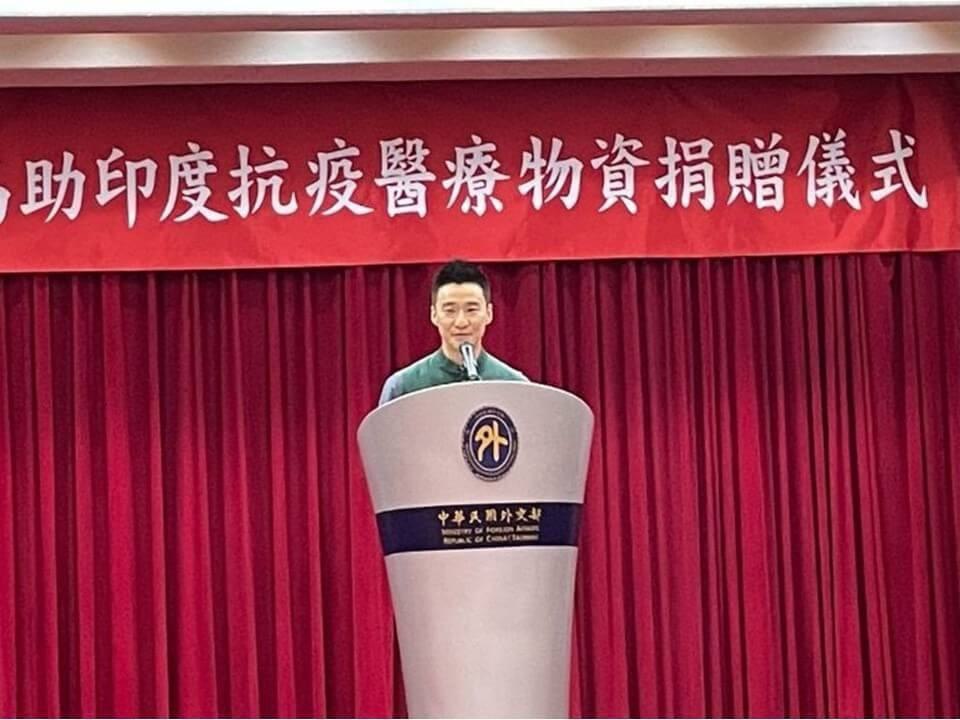}
        & \textcolor{ForestGreen}{Reference:} Chang Chien-Ting, an associate of the Adani Group, is a Chinese national. \newline\newline
        \textcolor{Mulberry}{MLLM+ICL:} Morris Chang, owner of PMC Projects, is the son of Chang Chung-ling, who is a director of Adani's companies and a business partner of Adani's brother, Vinod Adani. [1] \newline\newline
        \textcolor{MidnightBlue}{MICE:} Morris Chang, owner of PMC Projects, is the son of Chang Chung-ling, who is a director of Adani's companies and a business partner of Vinod Adani. [1] 
        & The LLM-based approaches fail to detect the deceptive framing and the nationality misattribution at the core of this post. This is likely because the post relies on background knowledge of India–China relations and domestic narratives around economic nationalism. The post falsely claims that the Adani Group’s (An Indian Multinational Conglomerate) associate is linked to a “Chinese” businessman, exploiting ongoing public suspicion toward Chinese influence in Indian infrastructure projects. Both MICE and MLLM+ICL extract the literal factual components (names, companies, and family relations) but fail to recognize the misleading intent: the deliberate conflation of Taiwan (“Republic of China”) with mainland China (“People’s Republic of China”) to frame the relationship as evidence of Chinese involvement. This illustrates a broader limitation of automated claim extraction, which in some cases would need context from real-world and real-time socio-political context.\\
    \hline
        This image is a fraud, created via AI. Supporters of russia, you go right ahead and believe this fraud. I have no objections at all. Supporters of Ukraine, Ukraine has taken at least five villages in the last day. The russian army in the last two days has lost immense amounts of men, far more than the daily average for this war. In the last village taken alone, fifty occupiers were slain, and four captured. Russian supporters, you won't believe me, and I absolutely do not need you to. What you believe or don't believe is immaterial. As we freed my family in Kherson, as we freed Kharkiv, so also now we free more villages from the horrible dystopia of the 'multi-polar paradise'. But you please do keep making your AI images and living in Putin's house of mirrors and illusions. \includegraphics[width=5cm]{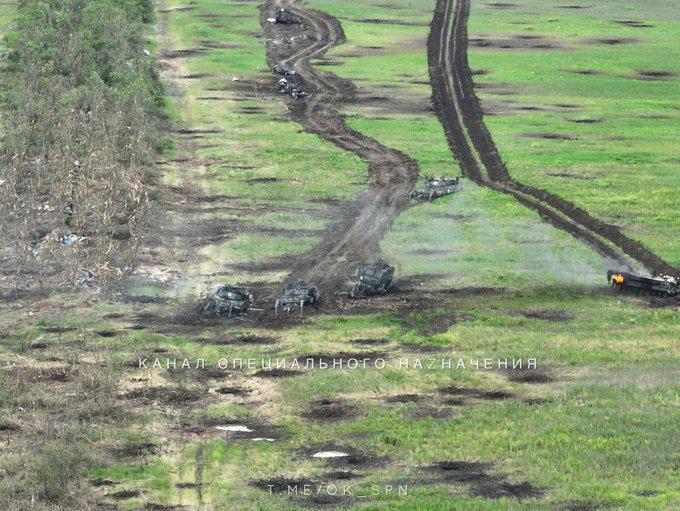}
        & \textcolor{ForestGreen}{Reference:} Image Showing Ukrainian Losses Is Generated By Artificial Intelligence. \newline\newline
        \textcolor{Mulberry}{MLLM+ICL:} Ukraine has taken at least five villages in the last day, and in the last village taken alone, fifty Russian soldiers were slain, and four captured. [1] \newline\newline
        \textcolor{MidnightBlue}{MICE:} Ukraine has taken at least five villages in the last day, and the Russian army has suffered immense losses in the last two days. [1] 
        & Both models extract a secondary claim from the lengthy post text but miss the primary, overarching claim that "This image is a fraud, created via AI" (the image is verified as legitimate). The models appear to latch onto the more concrete, detailed statements about recent military events later in the paragraph. This limitation likely arises from the claim extraction prompt, which directs the system to identify minimal or singular claims, a strategy that proves inadequate for posts containing longer, more complex paragraphs. \\
    \hline
\end{tabularx}
\caption{Analysis on instances where both baseline MLLMs and the MICE framework yields poor results, using Gemini 2.0 Flash as the underlying model. The numbers within the square brackets represent the LLM-based reference based score \textbf{(Part 2, continued from \ref{tab:error_analysis_part1}))}.}
\label{tab:error_analysis_part2}
\end{table*}

\begin{table*}[t]
\centering
\label{tab:smallerfont}
\footnotesize
\begin{tabularx}{\textwidth}{XXX}
    \hline
    \rowcolor[gray]{0.7}  
    \hline
    \multicolumn{3}{c}{\textit{\textbf{Cases where baseline MLLMs outperforms MICE}}} \\    
    \hline
    \textbf{Social Media Post} & \textbf{Claims} & \textbf{Analysis} \\
        \hline
        Tunisian Club African CEO being interviewed by a journalist about their game with Yanga SC \includegraphics[width=5cm]{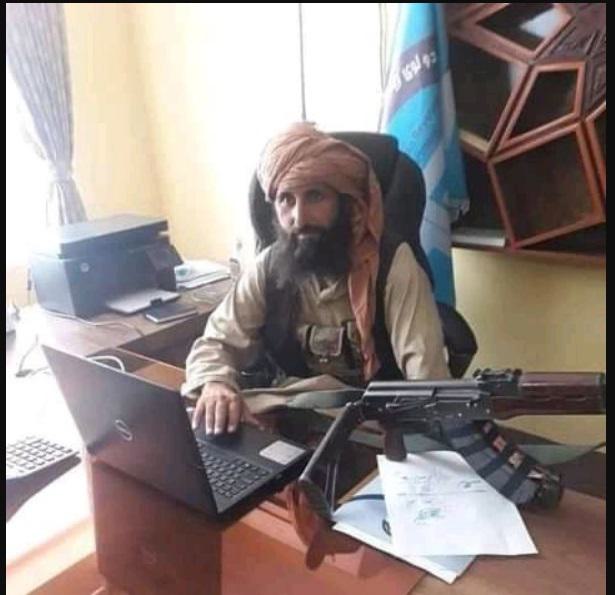}
        & \textcolor{ForestGreen}{Reference:} Image shows the CEO of Club Africain of Tunisia being interviewed after their match with Tanzania's Yanga SC. \newline\newline
        \textcolor{Mulberry}{MLLM+ICL:} The man in the photo is the CEO of Tunisian Club African being interviewed about their game with Yanga SC. [4]\newline\newline
        \textcolor{MidnightBlue}{MICE:} The CEO of Tunisian Club African is depicted as a man wearing a turban and traditional clothing, sitting at a desk with a laptop and an AK-47 assault rifle. [1]
        & This instance illustrates the potential drawback of applying MICE's complex reasoning to straightforward, literal posts. The MICE framework fails because of its over-emphasis on visual analysis where none is needed. The post's claim is fully contained in the text, but MICE's methodology compels it to integrate visual features. This shows that for simple, text-dominant claims, MICE's intent- and context-aware modules can introduce noise and errors rather than adding value. \\
    \hline
        Someone please ask Fashola how he got access to this server that has the election result data. INEC and APC rigged the election even before it started. They should come out and defend this. \includegraphics[width=5cm]{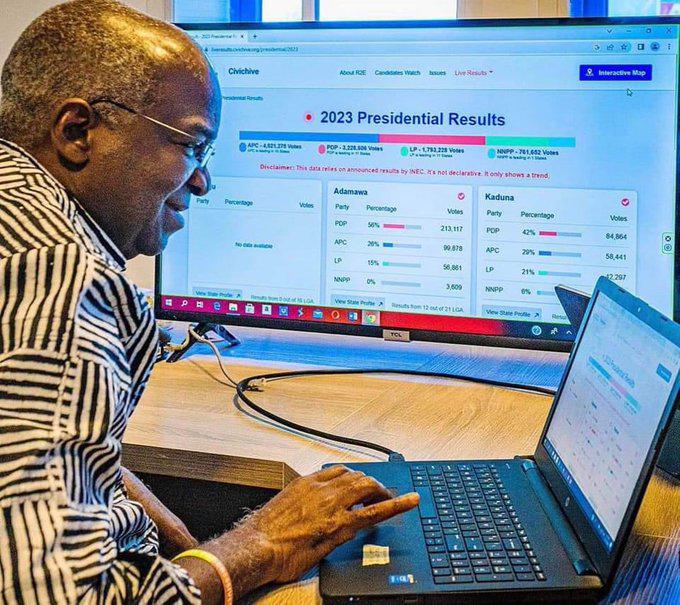}
        & \textcolor{ForestGreen}{Reference:} Nigerian housing minister Babatunde Fashola illegally accessed election data on a server prior to the official release of results. \newline\newline
        \textcolor{Mulberry}{MLLM+ICL:} A photo shows Babatunde Fashola accessing a server containing election result data. [4] \newline\newline
        \textcolor{MidnightBlue}{MICE:} The image shows election result data from Civichive for the 2023 Nigerian presidential election. [1]
        & MICE mentioned "result data from Civichive," a detail not mentioned in the text and likely derived from misinterpreting the website screenshot in the image. This shows that the Vision API or contextual breakdown module likely misinterpreted the UI elements in the image, whereas the baseline's more direct approach remained more faithful to the source text. \newline However, the MLLM+ICL baseline claim also does not capture the key detail that Fashola allegedly illegally accessed the server. The high score is arguably inflated, as the claim omits the critical detail from the reference that the access was "illegally" obtained. The automated judge rewards the partial semantic overlap, highlighting a weakness in the reference-based scoring metric itself. \\
    \hline
        U.S. District Judge Aileen Cannon (on the left side in photo) has been initially assigned to oversee the case...could preside over the trial as well. Cannon, appointed by Trump in 2019, made decisions that favored him in legal skirmishes during the documents investigation last year. Her rulings were overturned on appeal. Cannon would determine, among other things, when a trial would take place and what Trump's sentence would be if he were found guilty. \includegraphics[height=1em]{roll_eyes_emoji.png} \includegraphics[width=5cm]{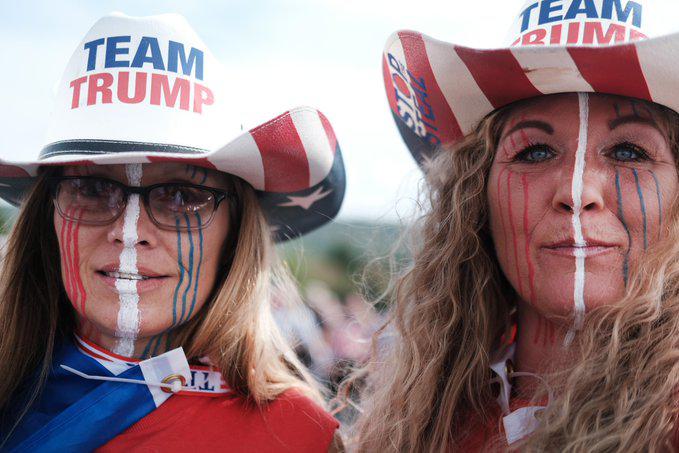}
        & \textcolor{ForestGreen}{Reference:} A photo shows Aileen Cannon, the US district court judge assigned to the criminal case against Donald Trump for retaining classified documents, wearing Trump apparel outside one of the former president's rallies. \newline\newline
        \textcolor{Mulberry}{MLLM+ICL:} Judge Aileen Cannon, who was appointed by Trump in 2019 and initially assigned to oversee his case, is shown in the photo wearing pro-Trump attire. [4] \newline\newline
        \textcolor{MidnightBlue}{MICE:} Aileen Cannon, a Trump appointee, has been assigned to oversee the classified documents case against Donald Trump, and she previously made decisions favorable to him that were later overturned. [2]
        & Here, the MLLM+ICL baseline outperforms MICE by successfully integrating key details from both the text and the image. The reference claim's core is that Judge Cannon is wearing "Trump apparel," a visual fact explicitly mentioned in the baseline's output. \newline In contrast, the MICE framework's treated the image as merely illustrative. Its reasoning process failed to grasp that the visual information was the most salient part of the claim. \\
    \hline
\end{tabularx}
\caption{Analysis on instances where the baseline MLLMs outperforms the MICE framework, using Gemini 2.0 Flash as the underlying model. The numbers within the square brackets represent the LLM-based reference based score.}
\label{tab:mice_failures}
\end{table*}

\end{document}